\documentclass[10pt,twocolumn,letterpaper]{article}

\usepackage[pagenumbers]{cvpr} 

\definecolor{cvprblue}{rgb}{0.21,0.49,0.74}
\usepackage[pagebackref,breaklinks,colorlinks,allcolors=cvprblue]{hyperref}

\def\paperID{2} 
\def\confName{CVPR}
\def\confYear{2025}

\title{Direction-Aware Hybrid Representation Learning for \\3D Hand Pose and Shape Estimation}

\author{
	Shiyong Liu\textsuperscript{1}\;\;\;
	Zhihao Li\textsuperscript{1}\;\;\;
        Xiao Tang\textsuperscript{1}\;\;\;
	Jianzhuang Liu\textsuperscript{2}\;\;\; \\
	{\textsuperscript{1}Huawei Noah's Ark Lab} \;\;\;\;{\textsuperscript{2}Shenzhen Institute of Advanced Technology}\;\;\;\; 
}

\begin{document}
\maketitle
\begin{abstract}
Most model-based 3D hand pose and shape estimation methods directly regress the parametric model parameters from an image to obtain 3D joints under weak supervision. However, these methods involve solving a complex optimization problem with many local minima, making training difficult. To address this challenge, we propose learning direction-aware hybrid features (DaHyF) that fuse implicit image features and explicit 2D joint coordinate features. This fusion is enhanced by the pixel direction information in the camera coordinate system to estimate pose, shape, and camera viewpoint. Our method directly predicts 3D hand poses with DaHyF representation and reduces jittering during motion capture using prediction confidence based on contrastive learning. We evaluate our method on the FreiHAND dataset and show that it outperforms existing state-of-the-art methods by more than 33\% in accuracy. DaHyF also achieves the top ranking on both the HO3Dv2 and HO3Dv3 leaderboards for the metric of Mean Joint Error (after scale and translation alignment). Compared to the second-best results, the largest improvement observed is 10\%. We also demonstrate its effectiveness in real-time motion capture scenarios with hand position variability, occlusion, and motion blur.
\end{abstract}    
\section{Introduction}
\label{sec:intro}

Estimating 3D hand pose and shape from monocular RGB images or videos is a critical research area in computer vision and graphics. It enables human-machine and human-environment interactions for various scenarios such as AR/VR, HCI and digital human. However, it is challenging due to depth ambiguity, motion blur, severe occlusion, low resolution, etc.

There are two categories of methods for 3D hand pose estimation: model-based and model-free. Model-based methods \cite{moon2022accurate, zimmermann2022contrastive, kulon2020weakly, zimmermann2019freihand} use parametric hand models as the proxy reoresentation. Model-free methods \cite{cai2018weakly, iqbal2018hand, spurr2020weakly, zimmermann2017learning, lin2021mesh} directly predict the 3D coordinates of hand joints. The advantage of the former over the latter is that they can use prior information from parametric models to avoid unreasonable prediction parameters. However, predicting 3D joint rotation from 2D RGB images is a highly non-linear process for neural networks, which causes pixel misalignment in model-based methods.

Some methods propose to involve 2D joint coordinates to improve 3D regression accuracy and 2D-3D consistency \cite{boukhayma20193d, Zhang_2021_ICCV, chen2021model}. However, most of them are not end-to-end because they require pre-training a 2D hand keypoint detector on a separate dataset. The accuracy of these methods depends heavily on the performance of the pre-trained detector and the quality of the dataset. Furthermore, their 2D joint features are mainly expressed as heatmaps which have large quantization errors in small and low-resolution targets. In order to reduce quantization errors, a regression-based end-to-end high-precision 2D keypoint detector that can be jointly optimized plays an important role in improving overall motion capture accuracy.

Directly regressing joint coordinates is a translation-dependent task. The CoordConv \cite{NEURIPS2018_60106888} method is proposed to deal with the translation invariance problem in the coordinate regression task with convolutional networks. CoordConv improves detection accuracy by adding corresponding channels to the input feature map of the convolution. These channels represent the coordinates of the pixels in the feature map, allowing the convolution learning process to perceive coordinates to some extent. However, Top-down 3D hand motion capture is a two-stage task. The hand is center-cropped and resized before being sent to the network. The coordinate encoding method of CoordConv cannot deal with 3D ambiguity. Encoding under the feature map coordinate system loses the hand position information in the full frame of the video, and does not improve the 3D estimation much.

A common problem in monocular video motion capture is the smoothness between frames. Since the hand targets have a relatively large range of motion in the entire frame, there may be problems with low resolution and motion blur. Time-domain filtering can effectively improve the smoothness between frames. However, the tracker may lose the target, resulting in poor-quality hand-image input to the network, which may cause false positives to the filter. A better idea is to get the confidence of each frame of the motion capture result. For those whose confidence is lower than a threshold, the result of high confidence in the previous frame can be used as the replacement. This can improve filter performance and alleviate the jittering and flipping problem.

Motivated by the above observations, we present an end-to-end representation learning with direction-aware hybrid features (DaHyF) to improve the accuracy of 3D hand pose and shape estimation. The direction-aware hybrid features are a fusion of implicit image features and explicit 2D joint coordinate features. To reduce quantization errors and obtain high accuracy 2D joint coordinates, we design an end-to-end sub-pixel coordinate predictor that can be jointly optimized. To alleviate the problem of weak spatial perception ability caused by the translation invariance of convolutional networks and to improve the model’s ability to regress 2D coordinates and 3D rotations, we develop a global direction map module. To avoid jittering and flipping caused by false positives, we propose a scheme of motion capture confidence calculation based on contrastive learning. The end-to-end training pipeline enables joint optimization using various datasets to improve the accuracy of each module and achieve more pixel-aligned 3D pose and shape estimation without an additional detector. Our method significantly outperforms current state-of-the-art methods on the FreiHAND \cite{zimmermann2019freihand}, HO3Dv2 \cite{hampali2020honnotate} and HO3Dv3 \cite{hampali2021ho} datasets according to various evaluation metrics.

Our main contributions are summarized as follows:
\begin{enumerate}
\item[$\bullet$]We design a 2D+3D end-to-end joint optimization framework with hybrid implicit image features and explicit joint coordinate features. Our sub-pixel coordinate classifier costs fewer resources and has smaller quantization errors compared to heatmap-based methods. The end-to-end pipeline allows us to jointly optimize 2D and 3D coordinates, building a complete contrastive learning strategy to generate prediction confidence and achieve mutual promotion between 2D and 3D pose estimation.
\item[$\bullet$]We propose a global direction map to enhance the spatial perception of convolutional networks. This module avoids the translation-agnostic problem and significantly improves both 2D and 3D pose estimation performance.
\item[$\bullet$]We present a motion capture confidence calculation scheme based on contrastive learning. It effectively utilizes non-hand patch information to reduce false positives and improve the robustness and smoothness in video motion capture.
\end{enumerate}
\section{Related Work}
\label{sec:Related Work}

\subsection{3D Hand Pose and Shape Estimation}
3D hand pose and shape estimation methods can be generally divided into two types: model-based and model-free. Model-based methods typically use a parameterized hand model \cite{romero2022embodied,ballan2012motion,khamis2015learning,tkach2016sphere} as a differentiable layer in a neural network to estimate shape and pose and map them to a triangle mesh and joint coordinates. Pose and shape are weakly supervised by supervising joint coordinates and the mask of rendered results. Since the parameterized model contains prior information about hand structure, this approach has the advantages of reducing the number of parameters, improving robustness, and reducing artifacts. However, there are also disadvantages: the parameters of the parameterized model are weakly supervised, resulting in difficulties in achieving pixel-aligned results, and the optimization process is more likely to fall into local minima \cite{li2019survey}. Model-free methods \cite{choi2020pose2mesh,moon2020i2l,kolotouros2019convolutional} do not require a predefined hand model and attempt to learn a mapping from the input image or depth data to pre-defined kinematic joints via an end-to-end network. They directly regress 2D/3D keypoint coordinates and calculate joint rotations using inverse kinematics. These methods generally achieve more pixel-aligned results compared to model-based methods, but they may suffer from noise and occlusion, resulting in artifacts due to the lack of hand prior information.

Our method combines the advantages of both approaches by additionally fusing coordinate features with image features to assist in regressing parameters of the parameterized hand model

\begin{figure*}[t]
\begin{center}
\includegraphics[width=0.93\textwidth]{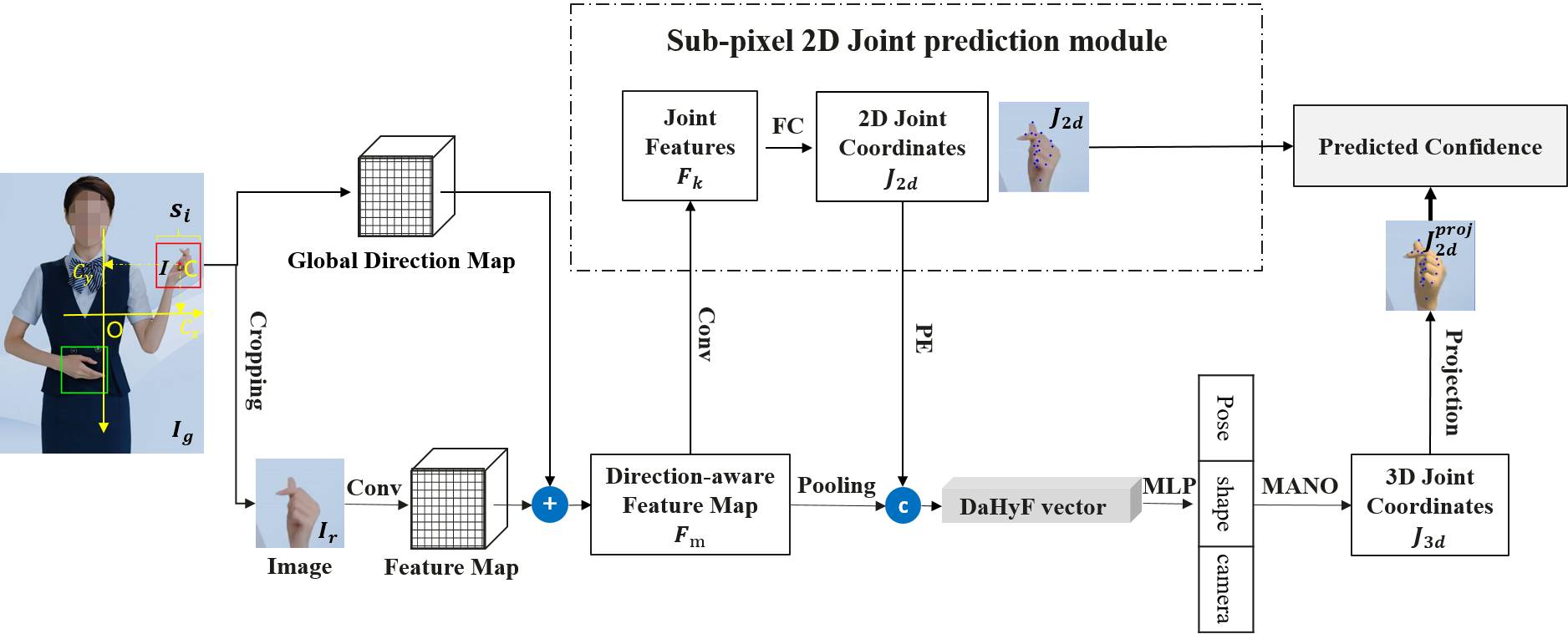}
\end{center}
   \caption{\textbf{Overview of DaHyF}. A hand region is cropped, resized and encoded as a local implicit feature map, which is then fused with a global direction map calculated from the hand bounding box. The fused features are used to detect 2D hand keypoints $J_{2d}$ in sub-pixel accuracy, which are positionally encoded (PE) with sinusoidal signals to build a joint feature vector. The pooled implict $F_m$ and the encoded explict $J_{2d}$ are concatenated to form our direction-aware hybrid feature (DaHyF) vector. This DaHyF vector is used to regress hand pose, shape, and camera parameters. Finally, 3D keypoints are obtained based on the MANO model \cite{romero2022embodied} and projected to 2D coordinates $J_{2d}^{proj}$ for confidence computation with $J_{2d}$.}
\label{fig:figure_whole_framework}
\end{figure*}
\subsection{2D Hand Keypoint Estimation}
2D hand keypoint estimation aims to locate the 2D coordinates of hand keypoints from images. This is usually done in a top-down manner because hand targets are generally small in the whole images. The top-down paradigm employs a two-step procedure that first detects hand bounding boxes and then performs single hand keypoint estimation for each bounding box. Top-down approaches can be categorized into regression-based \cite{toshev2014deeppose,qiu2020peeking,li2021pose} and heatmap-based \cite{sun2019deep,xiao2018simple,yuan2021hrformer,xu2022vitpose}. Regression-based methods directly regress the keypoint coordinates from the image, which are efficient and show promising potential in real-time applications, but they fail to provide a probability distribution for multiple candidate positions. To overcome the shortcomings of direct coordinate regression and make coordinates more suitable for regression by a convolutional neural network, heatmap-based joint representations have been proposed \cite{pfister2015flowing}, which output the keypoint positions as the peak values of heatmaps or confidence maps. The advantage of this type of methods is that it can provide a probability distribution for multiple candidate positions, but due to the quantization errors, heatmap-based methods do not perform well in low-resolution scenarios \cite{li2022simcc}, especially for small and severe motion blur targets such as hands.

Our method is regression-based and uses a sub-pixel coordinate classifier to reduce quantization errors.
\subsection{3D Pose Confidence}
Many works have proposed predicting pose confidence to promote the robustness and accuracy of networks. \cite{bridgeman2019multi} uses a 2D pose detector to provide a confidence value for each joint, with undetected joints having a confidence of zero.  \cite{tian2023multi} builds a multi-view 2D part confidence map to track 3D skeletons in the presence of missing detections, substantial occlusions, and large calibration errors. \cite{shotton2013real} generates confidence-scored 3D proposals for several body joints by reprojecting the classification result and finding local modes. \cite{gupta2019cullnet} targets the problem of inaccurate confidence values predicted by CNNs and takes pairs of pose masks rendered from a 3D model and crops regions in the original image as input to calibrate the confidence scores of the pose proposals. Most of these methods operate at the granularity of each joint, with confidence used to find the highest scoring result among many candidates or to judge the probability that the current joint is occluded.

In video motion capture, the accuracy of individual joints is important, but the overall hand pose has the greatest impact on the visual effect because it affects the smoothness of the motion capture process and whether there are sudden jittering or other abnormal mutations. Our method is designed to take into account all hand joints and output the confidence of the current frame’s motion capture result to alleviate jitter and flip caused by false positives, which can improve the robustness and smoothness of video motion capture.
\section{Method}
\label{sec:Method}
This paper presents a single-view 2D+3D end-to-end joint optimization framework with direction-aware hybrid features for hand motion capture. It provides comprehensive spatial information for the model-based approach and improves the pixel-alignment of hand motion capture. To achieve this, as shown in Figure \ref{fig:figure_whole_framework}, the framework has three components: (1) a global direction map, which encodes global direction information into an implicit image feature map to boost spatial perception capabilities of convolution layers, (2) sub-pixel joint coordinate classification, which predicts sub-pixel coordinate for each joint to reduce the quantization error, and (3) Positional Encoding (PE), which maps each coordinate into a high-dimensional space. The encoded 2D joint coordinates are fused with the implicit image features, which are then fed to the regressor to estimate MANO \cite{romero2022embodied} parameters. We also present a scheme for motion capture confidence calculation based on contrastive learning, which can improve the robustness and smoothness of the output video.

\begin{figure}[t]
\begin{center}
\includegraphics[width=1\linewidth]{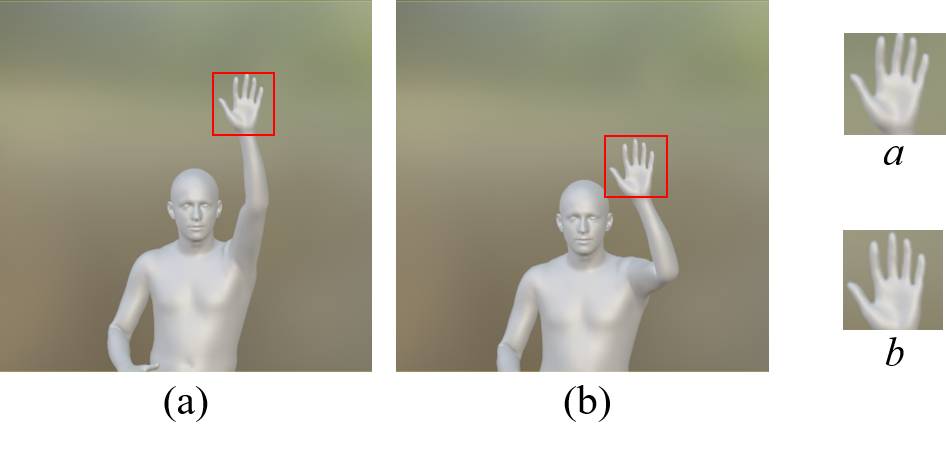}
\end{center}
   \caption{The input image features of $a$ and $b$ are very similar after cropping and resizing the hand patches. From their local direction maps, we can see that they are identical under the input image feature coordinate space. Thus, the local direction maps do not help CNN distinguish $a$ from $b$, and it is difficult to regress the global 3D pose information of the hands with respect to the camera coordinate system.}
\label{fig:local_pos_map2}
\end{figure}

\subsection{Hand Encoding}
Given an image or a video frame $I_g\in{R^{H\times W}}$ containing hands, we first crop square hand patches \emph{I $\in{R^{s_i\times s_i\times3}}$} and resize them to $s_p\times s_p$. Here, $s_i$ is the patch size and $s_p$ is the network input size equals to 224. We also flip all the left hand patches to the right. Then, we use convolution layers to extract shared low-level semantic features. Thus, the hand is represented by an implicit image feature map \emph{F}.
\subsection{Direction Map}
Convolutional neural networks (CNNs) are very successful in various visual tasks due to their translation invariance. However, this advantage becomes a defect in tasks that involve translation dependence, such as coordinate modeling, and potentially affects the final model performance \cite{NEURIPS2018_60106888}. One way to improve the accuracy of coordinate regression is to add a direction map to the feature map, which enables the latter convolution process to perceive the spatial information of the features. The traditional approach generally builds the direction map under the local coordinate space of image features. We define $s_f$ as the size of the image feature map $F$. The local direction map $L_m$ has the same size as $F$. Each channel contains either row coordinate $i\in{[0,s_f]}$ or column coordinate $j\in{[0,s_f]}$ under the local space of $F$. The origin is located at the upper left corner of $F$. The fused feature map is the concatenation of $F$ and $L_m$.

In our experiments, we find that the local direction encoding improves the estimation of 2D joints, but it does not enhance the 3D joint accuracy due to the ambiguity of the camera pose in 3D space. Compared with the body, the hand movement in the video is more flexible. As illustrated in Figure \ref{fig:local_pos_map2}, the hand can appear in different positions of the frame, after cropping and resizing, the caculated local direction maps are the same, and they cannot help CNN distinguish them. Therefore, it is hard to regress the 3D information of these hands with respect to the camera coordinate system.
\begin{figure*}[t]
\begin{center}
\includegraphics[width=0.81\linewidth]{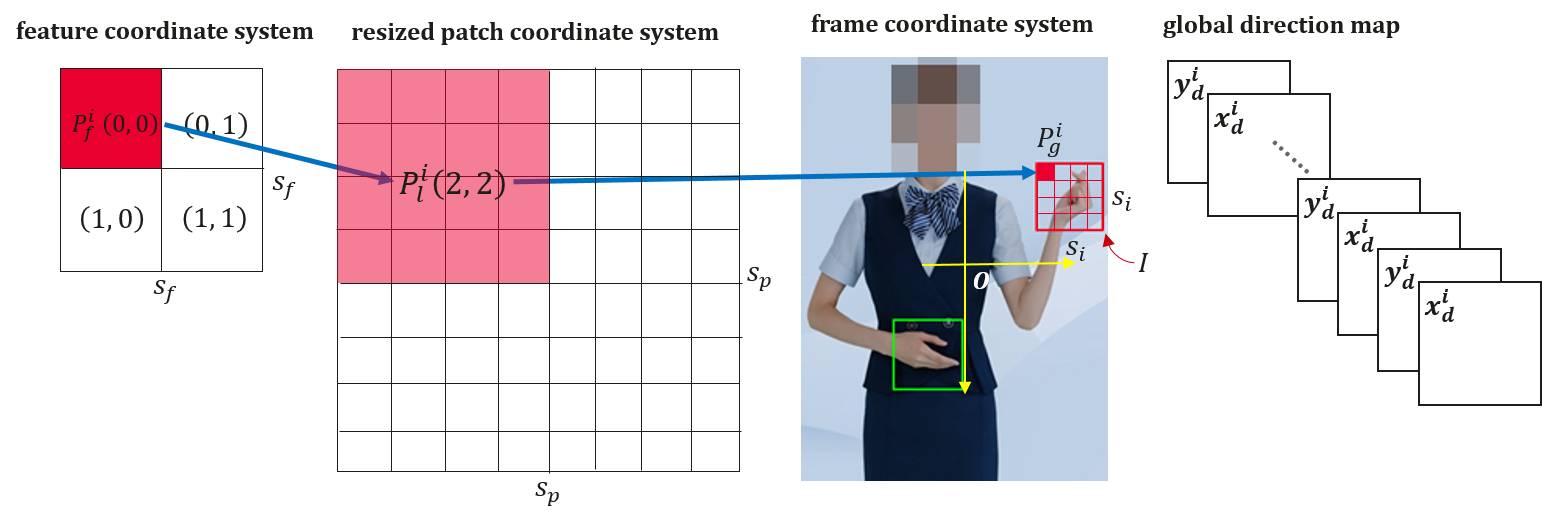}
\end{center}
   \caption{\textbf{Global direction map construction}. First, we calculate {$P_l^i$} from {$P_f^i$}, then {$P_g^i$} from {$P_l^i$} and {$P_d^i$} (not shown here) from {$P_g^i$}, and finally form the global feature map from {$P_d^i=(x_d^i, y_d^i)$}.}
\label{fig:global pos map}
\end{figure*}

To address this issue, we propose the global direction map, as illustrated in Figure \ref{fig:global pos map}. First, given a pixel coordinate $P_f^i$ in the image feature map, we calculate its corresponding pixel coordinate $P_l^i$ under the resized hand patch:
\begin{align}\label{eq2}
  {P_l^i} &= {P_f^i}\times {sc_p}+\frac{sc_p}{2},\\
  sc_p &= \frac{s_p}{s_f}.
\end{align}

Then, we calculate the corresponding global coordinate $P_g^i$ under the frame space:
\begin{align}\label{eq3}
  {P_g^i} &= {P_l^i}\times{sc_o}+{P_{ulc}}-O,\\
  sc_o &= \frac{s_i}{s_p}, \\
  O &= (\frac{W}{2}, \frac{H}{2}),
\end{align}
where $P_{ulc}$ is the upper left corner's coordinate of the hand patch $I$ in the global frame space, and $O$ is the origin located at the center of the frame.

To better estimate the joint rotation angle, we normalize the global coordinate $P_g^i$ into a direction vector $P_d^i$ in the camera coordinate system based on the focal length $f$, i.e., ${P_d^i} = {P_g^i} / {f}$, similar to the definition of the direction vector in NeRF \cite{mildenhall2021nerf}. In practice, if $f$ is unknown, we set it to $\sqrt{H^2+W^2}$, where $H$ and $W$ are the height and width of the frame.

Give the image feature map with all pixel coordinates $\{P_f^i\}$, we can obtain all their corresponding direction vectors $\{P_d^i\}$. Let $P_d^i=(x_d^i, y_d^i)$. We construct the global direction map as follows: We first form two channels: one with all the $\{x_d^i\}$ and the other with all the $\{y_d^i\}$. Then we copy them to form the global direction map such that it has the same channel number as that of the image feature map $F$ (see Figure \ref{fig:global pos map}).

\subsection{Joint Feature Fusion}
\textbf{2D Joint Estimation.}
The resolution of hand patches is usually low. This makes the traditional Gaussian heatmap-based methods unsuitable for coordinate regression due to quantization error. Therefore, we draw inspiration from SIMCC \cite{li2022simcc} and transform the coordinate estimation into a classification task. This can improve the localization accuracy of hand keypoints under low resolution. We apply a series of convolution operations on the image feature $F_m$ (see Figure \ref{fig:figure_whole_framework}) to obtain the representation $F_k\in{R^{21\times56\times56}}$ with 21 hand keypoints. Define $n_x$ (or $n_y$) as the number of class labels, where $n_x=s_p\times s$, $n_y=s_p\times s$, and $s\geq 1$ is a pre-defined scale for controlling the quantization error to improve the sub-pixel positioning accuracy ($s$ is set to 3 in this work). The numbers of bins for the horizontal and vertical axes are $n_x$ and $n_y$, respectively. The horizontal coordinate classifier and the vertical coordinate classifier predict the $k$-th keypoint $\{f_{x}^k, f_{y}^k\}$, where $k=1,2,...,21$, and $f_{x}^k, f_{y}^k \in{R^{nx}}$. Let the ground-truth labels be $\{\hat{J_{x}^k}, \hat{J_{y}^k}\}$. We train the network by smoothing the classification labels with 1D Gaussian and minimizing the KL divergence between $\{\overline{J}_{x}^k, \overline{J}_{y}^k\}$ and $\{f_{x}^k, f_{y}^k\}$, where $\{\overline{J}_{x}^k, \overline{J}_{y}^k\}$ is the smoothed result of $\{\hat{J}_{x}^k, \hat{J}_{y}^k\}$. Then we convert the network output $\{f_x^k, f_y^k\}$ to $J_{2d}\in{R^{2\times21}}$ through soft-argmax to obtain the predicted 2D keypoint coordinates: $J_{2d} = \{\frac{1}{s}\mathrm{\text{soft-argmax}}(f_x^k), \frac{1}{s}\mathrm{\text{soft-argmax}}(f_y^k))\}$.

\noindent\textbf{2D-3D Joint Features Fusion.}
To better combine the 2D joints with the direction-aware feature map $F_m$, we use a position encoding (PE) method similar to NeRF \cite{mildenhall2021nerf}, which maps the coordinates to a high dimensional space and enables our regressor to more easily approximate higher frequency information. The encoding result is as follows:
\begin{align}\label{eq4}
  \{\gamma_x^k{(\mu_{x}^k)}, \gamma_y^k{(\mu_{y}^k)}\},\quad \{\mu_{x}^k, \mu_{y}^k\} = \frac{(J_{2d}-\{\frac{s_p}{2}, \frac{s_p}{2}\})}{f},
\end{align}
\begin{multline}\label{eq4_1}
  \gamma_x^k{(p)} = \gamma_y^k{(p)} =  \\
  (sin(2^1\pi p),cos(2^1\pi p),...,sin(2^{L}\pi p),cos(2^L\pi p)),
\end{multline}
where $L$ is set to 4 in this work. So the encoded coordinates are $\{\gamma_x^k{(\mu_{x}^k)}, \gamma_y^k{(\mu_{y}^k)}\} \in {R^{2\times21\times (2L)}}$, which are then converted to a vector $V_{PE}\in{R^{336}}$. We also perform feature map pooling on $F_m$ to obtain another vector $V_{F_m}$. Finally, we concatenate $V_{PE}$ and $V_{F_m}$ to obtained the DaHyF feature vector (see Figure 1), which is fed to an MLP to regress the hand pose and shape, and camera parameters.

\subsection{Contrastive Learning for Pose Confidence}
The contrastive learning aims to reduce the impact of non-hand regions on hand pose estimation. We use the regressor (MLP in Figure \ref{fig:figure_whole_framework}) to generate the hand pose $\theta\in{R^{16\times 3}}$ (represented in axis angle), shape $\beta\in{R^{10}}$, and weak-perspective projection camera parameters $P_{weak}(s, tx, ty)\in{R^3}$ with respect to the cropped hand patch $I$. Then using MANO with $\theta$ and $\beta$ as the input, we obtain the 3D joint coordinates $J_{3d}$, which are projected to 2D joint coordinates $J_{2d}^{proj}$ by the global perspective projection parameters that are obtained based on $P_{weak}$ \cite{li2021pose}.

To measure the similarity between $J_{2d}$ and $J_{2d}^{Proj}$, we use cosine similarity as the contrastive learning criterion. We first normalize $J_{2d}$ and $J_{2d}^{Proj}$ by:
\begin{align}\label{eq6}
  \upsilon_{2d} &= (J_{2d}\times(\frac{s_i}{s_p}) - (\frac{s_i}{2}, \frac{s_i}{2}))/s_i, \\
  \upsilon_{2d}^{Proj} &= (J_{2d}^{Proj} - C)/s_i,
\end{align}
where $C$ is the center coordinate of the hand patch $I$ under the coordinate system of the frame $I_g$ (see Figure \ref{fig:figure_whole_framework}). Then all the coordinates in $\upsilon_{2d}$ (or $\upsilon_{2d}^{proj}$) are concatenated to form a vector  $\overline{\upsilon}_{2d}$ (or $\overline{\upsilon}_{2d}^{proj}$). Finally, $\overline{\upsilon}_{2d}$ and $\overline{\upsilon}_{2d}^{proj}$ are used to compute their cosine similarity.

\begin{figure}[t]
\begin{center}
\includegraphics[width=1\linewidth]{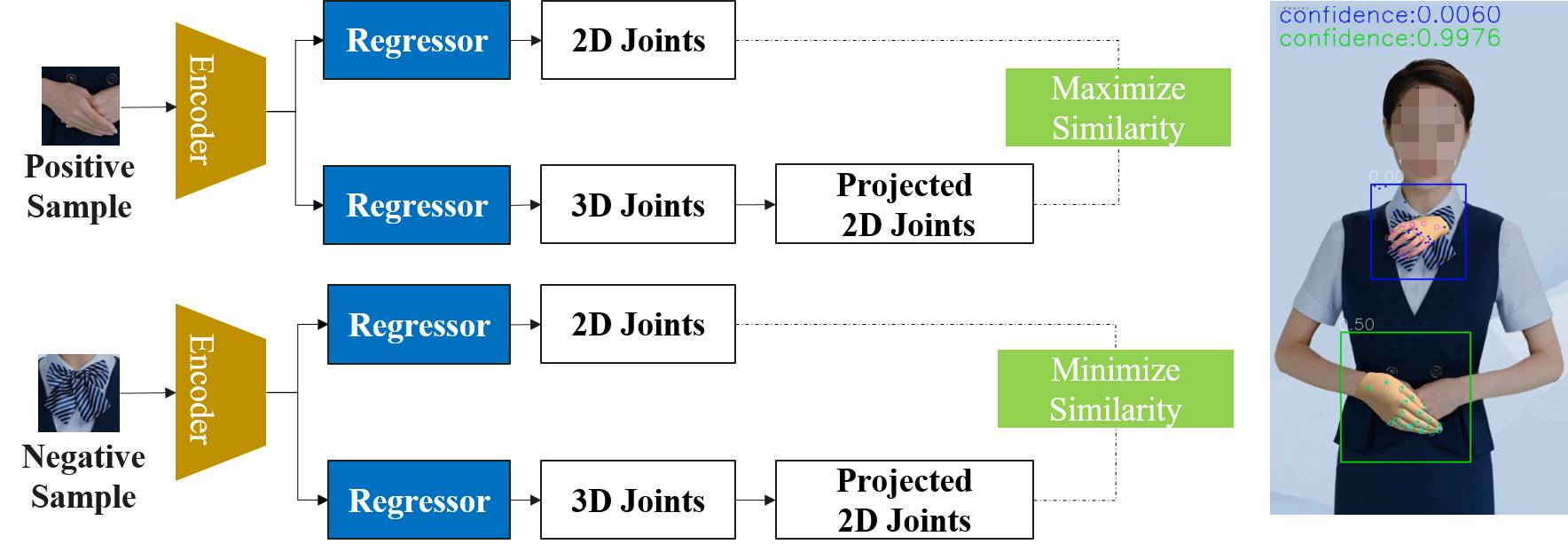}
\end{center}
   \caption{Our contrastive learning procedure.}
\label{fig:contrastive learning}
\end{figure}

When the patch $I$ contains the hand, $\overline{\upsilon}_{2d}$ and $\overline{\upsilon}_{2d}^{proj}$ are considered as a pair of positive samples. We also randomly crop patches from the frame, and when a patch does not contain the hand, the resulting $\overline{\upsilon}_{2d}$ and $\overline{\upsilon}_{2d}^{proj}$ are considered as a pair of negative samples. Our goal is to make a positive pair as close as possible and make a negative pair as far away as possible. Figure \ref{fig:contrastive learning} shows the main procedure of this contrastive learning. During inference, we use this cosine similarity value as the confidence measure for tidying the motion capture result. A patch whose confidence is lower than a threshold is considered as a false positive, and its hand pose parameters are replaced by those of its previous nearest hand patch.

\subsection{Training Objective}
The overall training loss, denoted as $\mathcal{L}$, consists of two components: $\mathcal{L}_{backbone}$ for backbone training and $\mathcal{L}_{c}$ for contrastive learning. The former includes 5 sub-losses: 2D branch loss ($\mathcal{L}_{2d}$), 3D loss ($\mathcal{L}_{3d}$), projected 2D loss ($\mathcal{L}_{2d}^{p}$), MANO loss ($\mathcal{L}_{m}$), and bone loss ($\mathcal{L}_{b}$).
\begin{equation}\label{eq8}
    \mathcal{L} = \mathcal{L}_{backbone} + \mathcal{L}_{c}.
\end{equation}
When 3D labels are available, we optionally apply $\mathcal{L}_{m}$ and $\mathcal{L}_{3d}$ and use the homoscedastic uncertainty strategy \cite{Zhang_2021_ICCV} to adaptively learn the weights of all sub-losses. $\mathcal{L}_{backbone}$ is defined as:
\begin{equation}\label{eq9}
  \mathcal{L}_{backbone} = \frac{\mathcal{L}_{2d}}{\sigma_{2d}^2}+\frac{\mathcal{L}_{3d}}{\sigma_{3d}^2}+\frac{\mathcal{L}_{2d}^p}{(\sigma_{2d}^p)^2}+\frac{\mathcal{L}_{m}}{\sigma_{m}^2}+\frac{\mathcal{L}_{b}}{\sigma_{b}^2},
\end{equation}
where $\sigma_{2d}$, $\sigma_{3d}$, $\sigma_{2d}^p$, $\sigma_{m}$, and $\sigma_{b}$ are learned by the strategy automatically. We use the L1 loss for $\mathcal{L}_{2d}$, $\mathcal{L}_{3d}$, and $\mathcal{L}_{2d}^{p}$, and the L2 loss for $\mathcal{L}_{m}$ and $\mathcal{L}_{c}$. The bone loss $\mathcal{L}_{b}$ refers to \cite{kulon2020weakly}.
\section{Experiments}
\subsection{Implementation Details}
We use \text{HRNet-W64} \cite{sun2019deep} as our backbone to extract the feature map $F$ from hand patches. The  backbone is initialized with ImageNet \cite{deng2009imagenet} pre-trained weights to leverage knowledge learned from large-scale image classification. Our model is trained with Adam optimizer \cite{kingma2014adam} on 4 GPUs with a batch size of 128. The hand detector is from \cite{lugaresi2019mediapipe} and a detected square hand patch is cropped and resized to $224\times 224$. We use an initial learning rate of $5\times10^{-5}$ and reduce it by a factor of 10 after 250 epochs out of 500 epochs. we only perform the contrastive learning in the last 100 epochs. To augment our training data, we apply random rotation, scaling, and cropping \cite{shorten2019survey}. These techniques increase the generalization and robustness of our model to different hand poses and orientations.

\subsection{Datasets and Evaluation Metrics}
We evaluate our method using three popular 3D hand reconstruction datasets: FreiHAND \cite{zimmermann2019freihand}, HO3Dv2 \cite{hampali2020honnotate} and HO3Dv3 \cite{hampali2021ho}. We achieve the top ranking on both the HO3Dv2 and HO3Dv3 leaderboards for the metric of Mean Joint Error (after scale and translation alignment).\footnote[1]{HO3Dv2: \href{https://codalab.lisn.upsaclay.fr/competitions/4318\#results}{https://codalab.lisn.upsaclay.fr/competitions/4318\#results}}$^{, }$ \footnote[2]{HO3Dv3: \href{https://codalab.lisn.upsaclay.fr/competitions/4393\#results}{https://codalab.lisn.upsaclay.fr/competitions/4393\#results}}

\noindent\textbf{FreiHAND}. FreiHAND \cite{zimmermann2019freihand} is a large-scale mixed real-world and synthetic dataset based on the MANO \cite{romero2022embodied} model. It contains 32560 training samples and 3960 test samples of people performing different hand movements. This dataset is suitable for evaluating the accuracy and realism of our method to reconstruct realistic hand meshes from RGB images.

\noindent\textbf{HO3D}. HO3Dv2 \cite{hampali2020honnotate} is a 3D hand-object dataset that contains 66,034 training samples and 11,524 evaluation samples. HO3Dv3 \cite{hampali2021ho} has more accurate annotations and more data including 83,325 training images and 20,137 testing images.  Evaluation on HO3Dv2 and HO3Dv3 are performed through an online submission website

\noindent\textbf{HanCo}. We also use HanCo \cite{zimmermann2022contrastive}, a dataset that contains 1518 short video clips captured by 8 calibrated and synchronized cameras, for training our model. It consists of 860304 individual frames and can be seen as an extended version of FreiHAND \cite{zimmermann2019freihand}. This dataset contains many non-hand frames, which are suitable for being negative samples for our contrastive learning.

\noindent\textbf{Evaluation Metrics}. We evaluate our method using 3D joint metrics and 2D joint metrics. For 3D joint metrics, we adopt procrustes-aligned mean per joint position error (PA-MPJPE), procrustes-aligned mean per vertex error (PA-MPVPE), and mean per joint position error (MPJPE). For conciseness, PA-MPJPE and PA-MPVPE are abbreviated as PJ and PV, respectively. They measure the Euclidean distances (in millimeter) of 3D joint or 3D mesh coordinates between the predictions and ground truth \cite{zhou2018monocap, kanazawa2018end, kolotouros2019learning,kolotouros2019convolutional}. Additionally, we calculate the F-Score at specific distance thresholds, denoted as F@5 and F@15, which correspond to thresholds of 5mm and 15mm, respectively. This score represents the harmonic mean of recall and precision between two meshes with respect to the given threshold \cite{chen2022mobrecon}. For 2D joint metrics, we adopt the average endpoint error (EPE) \cite{spurr2021self} in pixels. 

\subsection{Comparison with State-of-the-Art Methods}
We compare our method with current state-of-the-art (SOTA) works on 3D hand pose estimation. We use the CLIFF annotator \cite{li2022cliff} to provide MANO \cite{romero2022embodied} pseudo-GT for the FreiHAND \cite{zimmermann2019freihand} dataset. This allows us to train our model with more realistic and accuracy hand shapes and poses. We evaluate our method against model-based \cite{zimmermann2019freihand, feng2021collaborative, rong2021frankmocap, tang2021towards, zhang2021pymaf, chen2022mobrecon, li2022cliff, pang2024towards, chen2021model, jiang2023probabilistic, moon2022accurate, kulon2020weakly} and model-free \cite{lin2021end, lin2021mesh, moon2020i2l, choi2020pose2mesh} approaches on FreiHAND. As shown in Table \ref{sota comparison}, our method achieves significant improvements over the current SOTA Mesh Graphormer \cite{lin2021mesh} in all metrics. This demonstrates the effectiveness of our method for 3D hand pose estimation.

As shown in Table \ref{sota comparison_ho3dv2} and Table \ref{sota comparison_ho3dv3}, we also conduct evaluation on the HO3Dv2 \cite{hampali2020honnotate} and HO3Dv3 \cite{hampali2021ho} datasets, which are more challenging than FreiHAND \cite{zimmermann2019freihand} due to severe object occlusion. Our DaHyF, trained on Freihand \cite{zimmermann2019freihand} and with only three additional fine-tuning epochs on each of the HO3D datasets, achieves the first place on both the leaderboards for the Mean Joint Error metric (after scale and translation alignment), demonstrating its excellent generalizability.
\begin{table}[t]
\caption{Performance comparison with SOTA methods on the FreiHAND \cite{zimmermann2019freihand} test set.}\label{sota comparison}
\begin{center}
\resizebox{0.48\textwidth}{!}{
\begin{tabular}[1.0\textwidth]{lccccc}
\hline
Method & Venue & PJ $\downarrow$ & PV $\downarrow$ & F@5 $\uparrow$ & F@15 $\uparrow$ \\
\hline
\multicolumn{5}{l}{\textbf{* Model-free}} \\
Graphormer \cite{lin2021mesh} & ICCV'21 & 5.9 & 6.0 & 76.4 & 98.6\\
METRO \cite{lin2021end} & CVPR'21 & 6.3 & 6.5 & 73.1 & 98.4 \\
I2L-MeshNet \cite{moon2020i2l} & ECCV'20 & 7.4 & 7.6 & 68.1 & 97.3 \\
Pose2Mesh \cite{choi2020pose2mesh} & ECCV'20 & 7.7 & 7.8 & 67.4 & 96.9\\
\hline
\multicolumn{5}{l}{\textbf{* Model-based}} \\
MANO CNN \cite{zimmermann2019freihand} & ICCV'19 & 11.0 & 10.9 & 51.6 & 93.4\\
FrankMocap \cite{rong2021frankmocap} & ICCV'21 & 9.2 & 11.6 & 55.3 & 95.1\\
PIXIE \cite{feng2021collaborative} & 3DV'21 & 12.2 & 11.8 & 46.8 & 91.9 \\
Tang et al. \cite{tang2021towards} & CVPR'21 & 6.7 & 6.7 & 72.4 & 98.1 \\
Moon et al. \cite{kulon2020weakly} & CVPR'20 & 8.4 & 8.6 & 61.4 & 96.6  \\
Hand4Whole \cite{moon2022accurate} & CVPR'22 & 7.7 & 7.7 & 66.4 & 97.1 \\
CLIFF \cite{li2022cliff} & ECCV'22 & 6.8 & 6.6 & - & - \\
PyMAF \cite{zhang2021pymaf} & CVPR'21 & 7.5 & 7.7 & 67.1 & 97.4 \\
MobRecon \cite{chen2022mobrecon} & CVPR'22 & 6.9 & 7.2 & 69.4 & 97.9 \\
RoboSMPLX \cite{pang2024towards} & NIPS'24 & 6.9 & 6.7 & 71.5 & 98.1 \\
S$^2$Hand \cite{chen2021model} & CVPR'21 & 11.8 & 11.9 & 48.0 & - \\
AMVUR \cite{jiang2023probabilistic} & CVPR'23 & 6.2 & 6.1 & 76.7 & 98.7 \\
HaMeR \cite{pavlakos2024reconstructing} & CVPR'24 & 6.0 & 5.7 & 78.5 & 99.0 \\
\hline
DaHyF (Ours) & & \textbf{4.0} & \textbf{4.7} & \textbf{84.8} & \textbf{99.8} \\
\hline
\end{tabular}}
\end{center}
\end{table}

\begin{table}[t]
\caption{Performance comparison with SOTA methods on the HO3Dv2 test set.}\label{sota comparison_ho3dv2}
\begin{center}
\resizebox{0.48\textwidth}{!}{
\begin{tabular}[1.0\textwidth]{lccccc}
\hline
Method & Venue & PJ $\downarrow$ & PV $\downarrow$ & F@5 $\uparrow$ & F@15 $\uparrow$ \\
\hline
ObMan \cite{hasson2019learning} & CVPR'19 & 11.0 & 11.0 & 46.4 & 93.9 \\
HO3D \cite{hampali2020honnotate} & CVPR'20 & 10.7 & 10.6 & 50.6 & 94.2 \\
I2UV-HandNet \cite{chen2021i2uv} & CVPR'21 & 9.9 & 10.1 & 50.0 & 94.3 \\
MobRecon \cite{chen2022mobrecon} & CVPR'22 & 9.2 & 9.4 & 53.8 & 95.7 \\
METRO \cite{lin2021end} & CVPR'21 & 10.4 & 11.1 & 48.4 & 94.6 \\
S$^2$Hand \cite{chen2021model} & CVPR'21 & 11.4 & 11.2 & 45.0 & 93.0 \\
AMVUR \cite{jiang2023probabilistic} & CVPR'23 & 8.3 & 8.2 & 60.8 & 96.5 \\
\hline
DaHyF (Ours) & & \textbf{8.0} & \textbf{8.1} & \textbf{61.2} & \textbf{97.6}\\
\hline
\end{tabular}
}
\end{center}
\end{table}

\begin{table}[t]
\caption{Performance comparison with SOTA methods on the HO3Dv3 test set.}\label{sota comparison_ho3dv3}
\begin{center}
\resizebox{0.48\textwidth}{!}{
\begin{tabular}[1.0\textwidth]{lccccc}
\hline
Method & Venue & PJ $\downarrow$ & PV $\downarrow$ & F@5 $\uparrow$ & F@15 $\uparrow$ \\
\hline
ArtiBoost \cite{yang2022artiboost} & CVPR'22 & 10.8 & 10.4 & 50.7 & 94.6 \\
Keypoint Trans \cite{hampali2022keypoint} & CVPR'22 & 10.9 & - & - & - \\
AMVUR \cite{jiang2023probabilistic} & CVPR'23 & 8.7 & 8.3 & 59.3 & 96.4 \\
S$^2$Hand \cite{chen2021model} & CVPR'21 & 11.5 & 11.1 & 44.8 & 93.2 \\
\hline
DaHyF (Ours) & & \textbf{7.5} & \textbf{7.5} & \textbf{63.7} & \textbf{97.4}\\
\hline
\end{tabular}}
\end{center}
\end{table}

\begin{figure*}[t]
\begin{center}
\includegraphics[width=\linewidth]{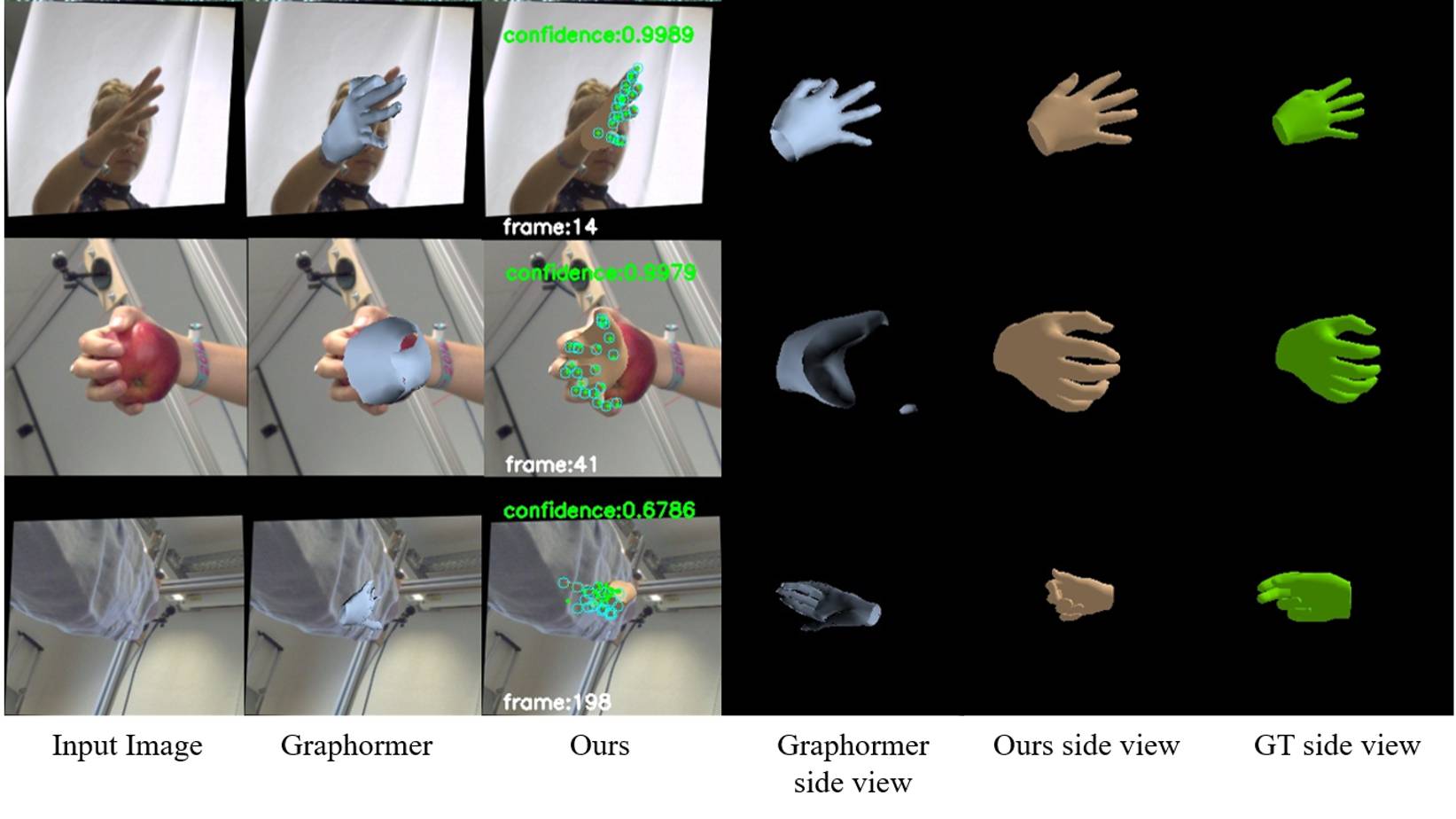}
\end{center}
   \caption{Qualitative comparison on the FreiHAND test set between our model and Mesh Graphormer. The last three columns visualise the results from novel views. Our results have higher accuracy and fewer artifacts in occluded scenes. Our model also provides confident scores for the pose estimation.}
   \label{fig:comparision between Mesh Graphormer}
\end{figure*}

Figure \ref{fig:comparision between Mesh Graphormer} shows qualitative results obtained by our method and Mesh Graphormer \cite{lin2021mesh} on the FreiHAND test set, where the reconstructed hand meshes are rendered. We can see that our results are much more accurate and pixel-aligned with the hands in the images.
\begin{table*}[t]
\caption{Ablation studies of the components of our model. LDM: local direction map. GDM: global direction map. Heatmap: heatmap-based joint coordinate regression. SJCP: sub-pixel joint 2D coordinate prediction module. PE: position encoding. $\mathcal{L}_{c}$: contrastive learning loss.}\label{ablation_study}
\begin{center}
\small\begin{tabular}[1\textwidth]{lcccc}
\hline
& PA-MPJPE $\downarrow$ & MPJPE $\downarrow$ & EPE ($J_{{2d}}^{{proj}}$) $\downarrow$ & EPE ($J_{2d}$) $\downarrow$ \\
\hline
Baseline & 8.91 & 18.14 & 7.37 & -\\
Baseline + GDM & 8.83 & 17.83 & 7.01 & - \\
Baseline + GDM + Heatmap & 8.40 & 17.30 & 6.80 &8.10 \\
Baseline + GDM + Heatmap + $\mathcal{L}_{c}$ & 8.20 & 16.99 & 6.33 & 7.36 \\
Baseline + GDM + SJCP + $\mathcal{L}_{c}$ & 8.02 & 16.11 & 6.17 & \textbf{6.95} \\
Baseline + LDM + SJCP + $\mathcal{L}_{c}$ & 9.59 & 22.7 & 10.03 & 9.06 \\
\hline
Baseline + GDM + SJCP + PE + $\mathcal{L}_{c}$ (Ours) & \textbf{7.88} & \textbf{15.95} & \textbf{6.05} & 7.83 \\
\hline
\end{tabular}
\end{center}
\end{table*}

\subsection{Ablation Study}
To evaluate the effectiveness of our proposed framework, we conduct an ablation study using the FreiHAND dataset. Initially, we establish a Baseline model, which solely comprises the lower branch as depicted in Figure 1, employing ResNet-50 as the backbone \cite{he2016identity}. Subsequently, we incrementally incorporate different components into the Baseline model to construct five distinct models. The qualitative results obtained from these experiments are summarized in Table \ref{ablation_study}.

These results reveal several key findings. Firstly, the integration of components such as GDM (Global Direction Map), $\mathcal{L}_{c}$ (contrastive learning loss), SJCP (Sub-Pixel Joint Coordinate Prediction Module), and PE (Position Encoding) significantly enhances the performance of the Baseline model. Particularly noteworthy is the observation that the inclusion of the proposed GDM component leads to notable performance improvements across all evaluation metrics when compared with the Baseline+LDM+SJCP (+ $\mathcal{L}_{c}$) and Baseline+GDM+SJCP (+ $\mathcal{L}_{c}$) configurations.

Furthermore, the comparison between the Baseline+GDM+Heatmap  (+ $\mathcal{L}_{c}$) and Baseline+GDM+SJCP  (+ $\mathcal{L}_{c}$) configurations highlights the superiority of SJCP over heatmap-based joint coordinate regression methods. This indicates the effectiveness of our sub-pixel joint coordinate classification approach in accurately predicting joint coordinates.

Ultimately, our final model, denoted as Baseline+GDM+SJCP+PE  (+ $\mathcal{L}_{c}$), emerges as the top-performing configuration, demonstrating superior performance across all evaluated metrics. These findings underscore the efficacy of our proposed framework in enhancing the accuracy and robustness of hand motion capture systems.
\section{Limitation and Conclusion}
\label{sec:Limitation and Conclusion}

\textbf{Limitation}. Our method only considers single-hand movement without occlusions by two hands. If the hand detector fails to distinguish between left and right hands, our model may give wrong results.

\noindent\textbf{Conclusion}. We have presented a novel single-view 2D+3D end-to-end joint optimization framework augmented with direction-aware hybrid features, aimed at enhancing the accuracy of hand motion capture. These direction-aware hybrid features are a blend of implicit image features and explicit 2D joint coordinate features, providing a comprehensive representation of hand motion.

To mitigate issues such as jittering and flipping induced by false positives, we have proposed a motion capture confidence calculation scheme based on contrastive learning. This approach helps enhance the robustness of our model against erroneous detections.

Experiments on the FreiHAND dataset demonstrate the effectiveness of our method, revealing a significant improvement of more than 33\% in accuracy compared to existing state-of-the-art techniques. Our model also achieves top ranking on both the HO3Dv2 and HO3Dv3 leaderboards for the metric of Mean Joint Error. These results underscore the potential of our framework to advance the field of hand motion capture, offering enhanced performance and reliability in real-world applications.
{
    \small
    \bibliographystyle{ieeenat_fullname}
    \bibliography{main}
}

\clearpage
\setcounter{page}{1}
\maketitlesupplementary

\setcounter{figure}{0}
\setcounter{section}{0}
\renewcommand{\thesection}{\Alph{section}}

In this supplementary material, we report the Percentage of Correct Keypoints (PCK) metric for 3D keypoints on the FreiHAND and HO3Dv2 datasets, provide qualitative results on two additional datasets (COCO \cite{lin2014microsoft} and RHD \cite{zimmermann2017learning}), along with three demo videos, and finally show some failure cases.

\section{PCK on FreiHAND and HO3Dv2}
To quantitatively assess the performance of the proposed DaHyF, we conduct an extensive evaluation using the PCK metric. The PCK metric is determined by normalizing the Euclidean distances between the predicted and Ground Truth (GT) keypoints by the length of the head segment. A keypoint is deemed correct if its normalized distance falls below a predefined threshold. We plot the PCK curves for thresholds ranging from 0cm to 5cm.

\begin{figure}[t]
\begin{center}
\includegraphics[width=0.91\linewidth]{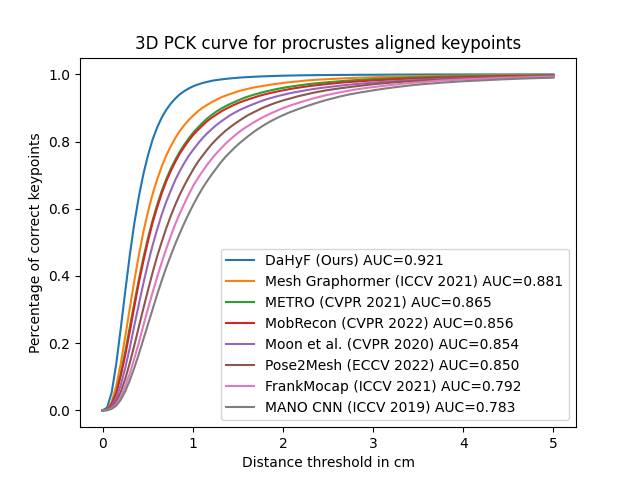}
\end{center}
   \caption{3D PCK on FreiHAND.}
   \label{fig:freihand_pck}
\end{figure}

\begin{figure}[t]
\begin{center}
\includegraphics[width=0.91\linewidth]{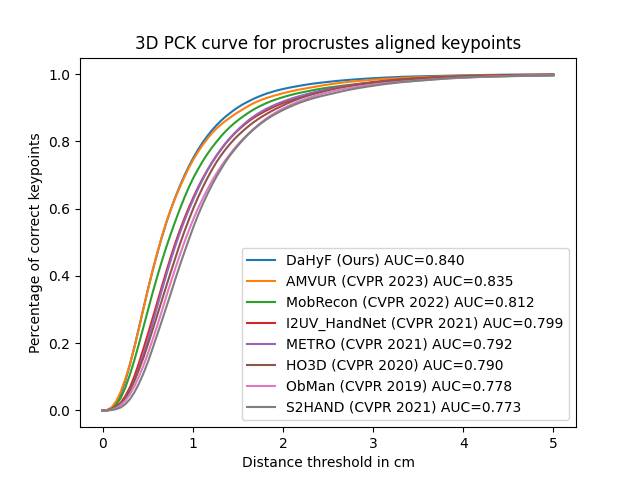}
\end{center}
   \caption{3D PCK on HO3Dv2.}
   \label{fig:ho3dv2_pck}
\end{figure}

As illustrated in Figures \ref{fig:freihand_pck} and \ref{fig:ho3dv2_pck}, our DaHyF exhibits superior performance in the 3D PCK metric for procrustes aligned keypoints.

\section{Quantitative Results on RHD and COCO}
\begin{figure*}[t]
\begin{center}
\includegraphics[width=\linewidth]{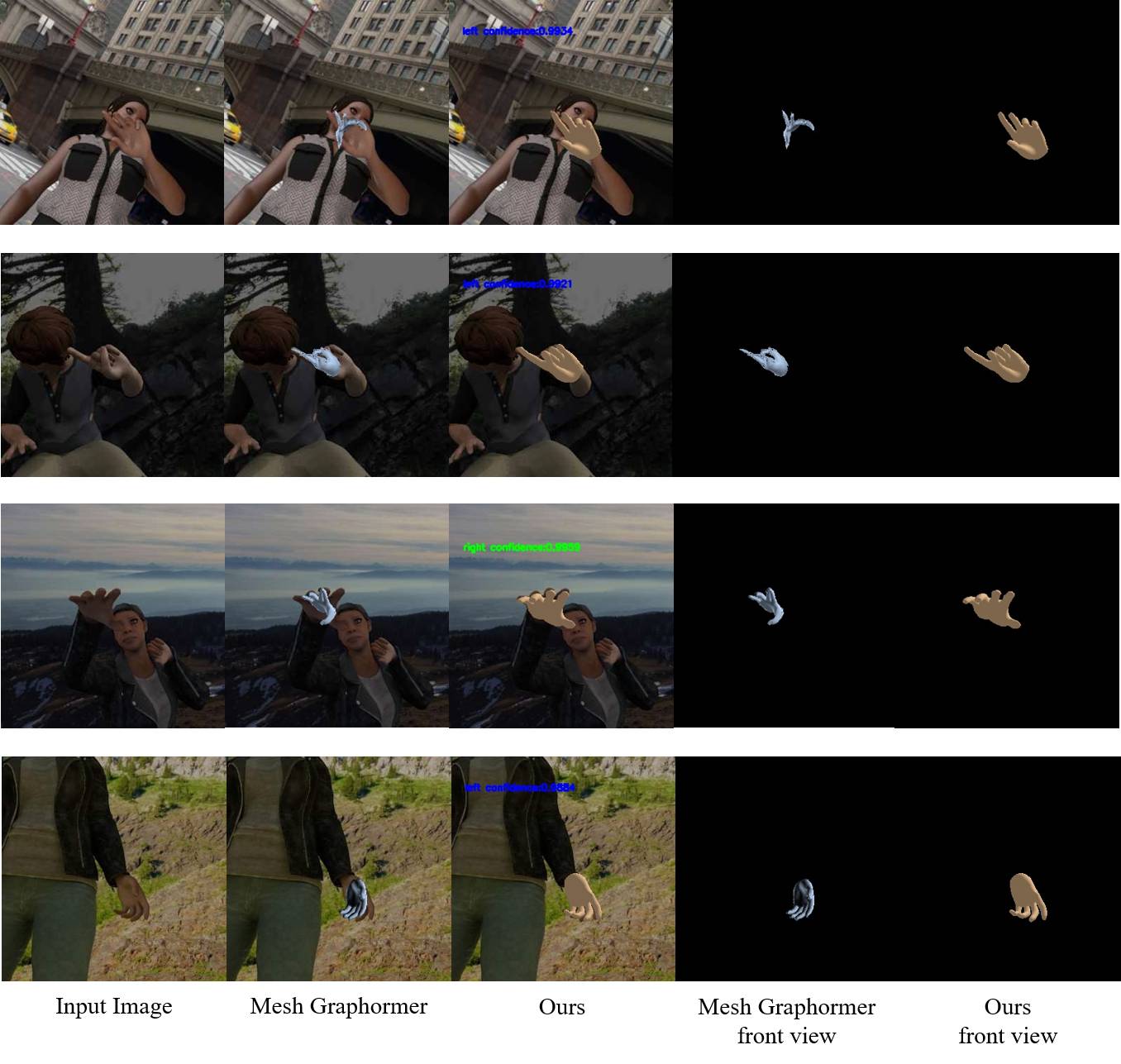}
\end{center}
   \caption{Qualitative comparison between DaHyF and Mesh Graphormer on the RHD dataset.}
   \label{fig:RHD dataset}
\end{figure*}

\begin{figure*}[t]
\begin{center}
\includegraphics[width=0.9\linewidth]{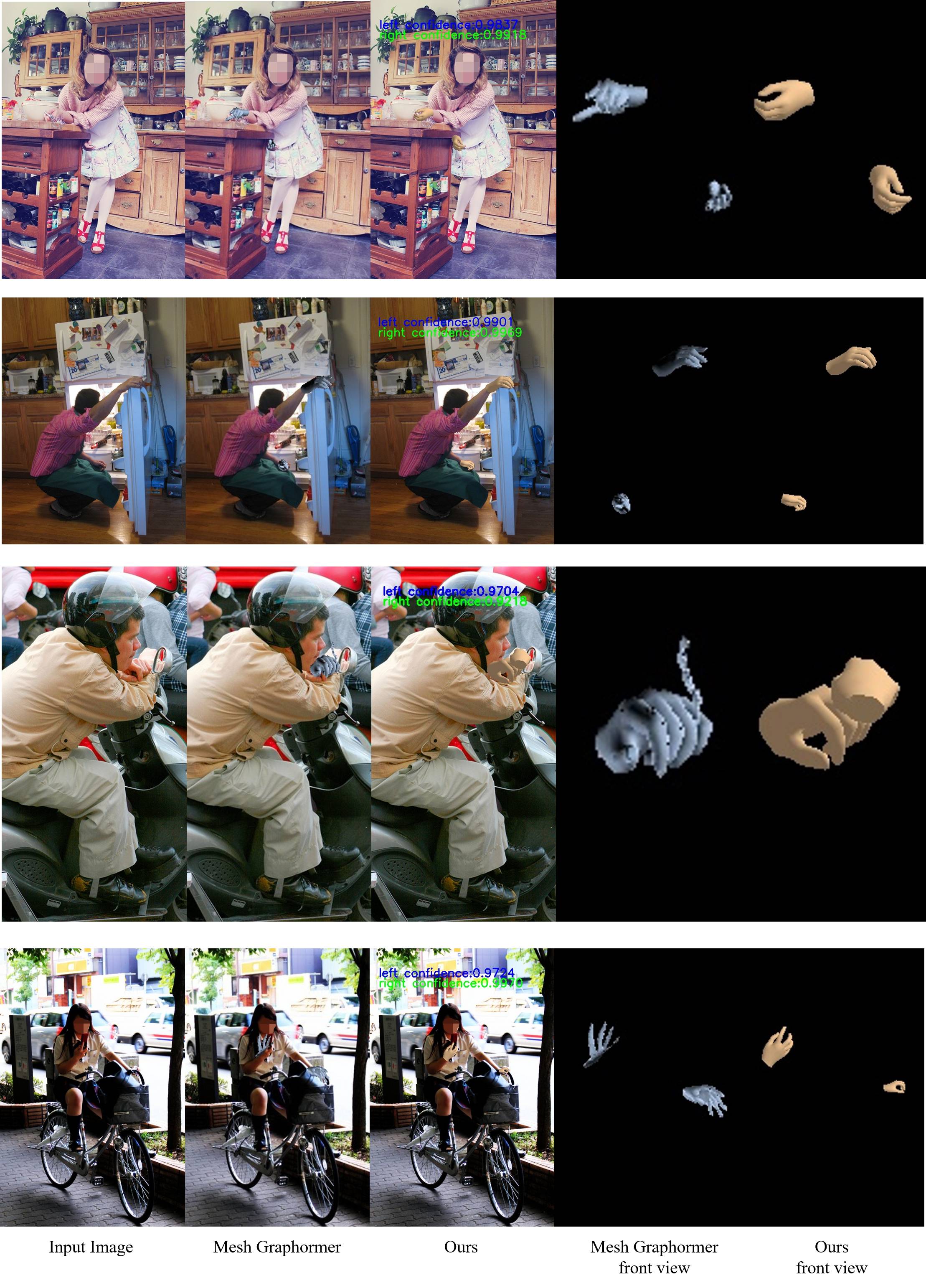}
\end{center}
   \caption{Qualitative comparison between DaHyF and Mesh Graphormer on the COCO dataset .}
   \label{fig:COCO dataset}
\end{figure*}

In Figures \ref{fig:RHD dataset} and \ref{fig:COCO dataset}, we compare
our method with Mesh Graphormer (one of the state-of-the-art methods) on the RHD and COCO datasets, respectively. Both use HRNet-W64 as the backbone.
As shown in the figures, DaHyF is effective at handling challenging cases such as small hands, occlusion, and appearance variance, thanks to our proposed 2D+3D end-to-end joint optimization framework with direction-aware hybrid features.

\section{Video Results on Temporal Filtering with the Predicted Confidence}
\begin{figure*}[t]
\begin{center}
\includegraphics[width=\linewidth]{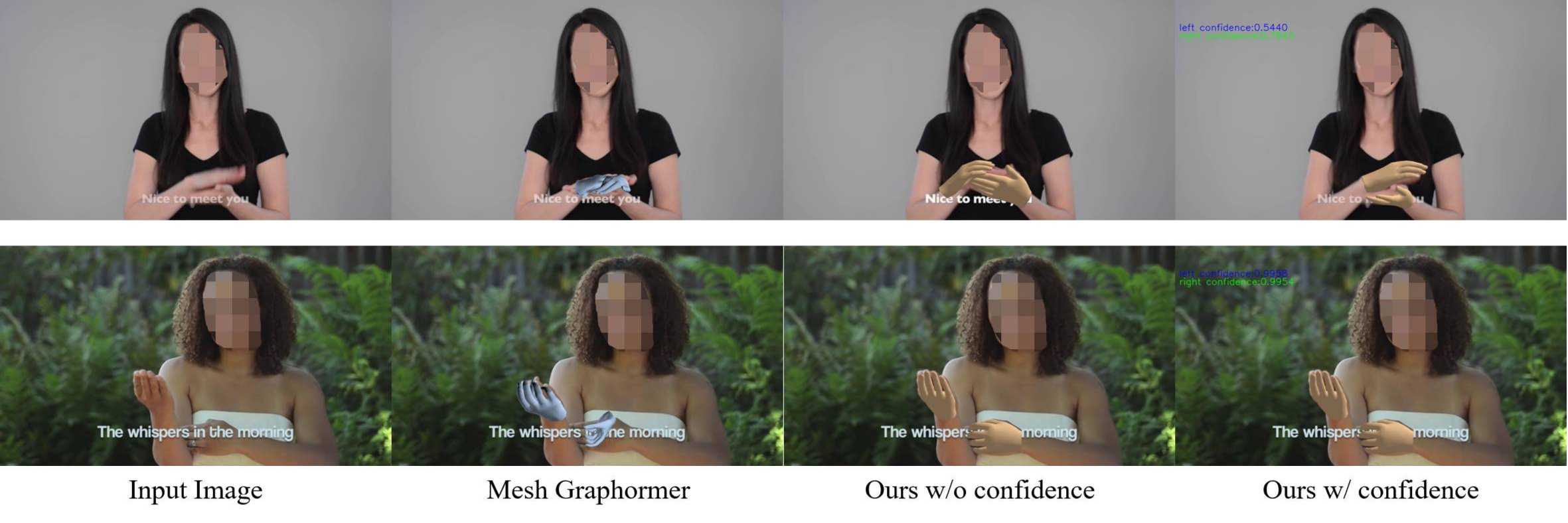}
\end{center}
   \caption{Qualitative results on three test videos.}
   \label{fig:video result}
\end{figure*}

We demonstrate the effectiveness of our confidence prediction module on three videos with small hands and severe motion blur. Figure \ref{fig:video result} shows one extracted frame from each of the three videos. Our method provides more robust and smoother results with less jittering and flipping, thanks to the predicted confidence in temporal filtering. The three videos are also provided separately in the supplementary material: 
\begin{enumerate}
\item[$\bullet$]Video\_Comparison\_with\_Mesh\_Graphormer\_1.mp4
\item[$\bullet$]Video\_Comparison\_with\_Mesh\_Graphormer\_2.mp4
\item[$\bullet$]Video\_Comparison\_with\_Mesh\_Graphormer\_3.mp4
\end{enumerate}

\section{Failure Cases}
\begin{figure*}[t]
\begin{center}
\includegraphics[width=\linewidth]{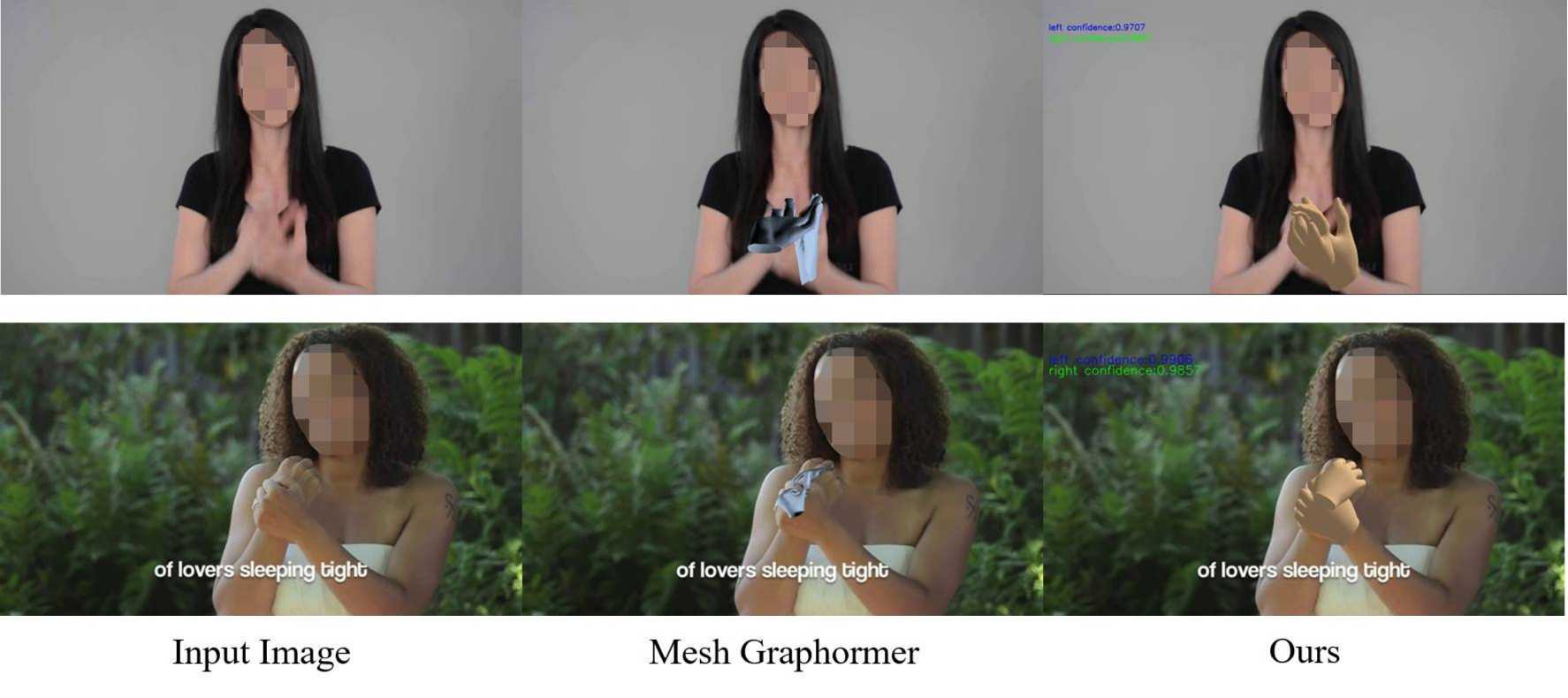}
\end{center}
   \caption{Failure cases.}
   \label{fig:failure cases}
\end{figure*}

We show some examples in Figure \ref{fig:failure cases} where our method fails to estimate the poses correctly. When there is severe occlusion or motion blur across multiple consecutive frames, our method faces challenges in accurately estimating the hand poses. Note that Mesh Graphormer also fails to do so.


\end{document}